\documentclass[letterpaper]{article} % DO NOT CHANGE THIS
\usepackage{aaai24}  % DO NOT CHANGE THIS
\usepackage{times}  % DO NOT CHANGE THIS
\usepackage{helvet}  % DO NOT CHANGE THIS
\usepackage{courier}  % DO NOT CHANGE THIS
\usepackage[hyphens]{url}  % DO NOT CHANGE THIS
\usepackage{graphicx} % DO NOT CHANGE THIS
\usepackage{natbib}  % DO NOT CHANGE THIS AND DO NOT ADD ANY OPTIONS TO IT
\usepackage{caption} % DO NOT CHANGE THIS AND DO NOT ADD ANY OPTIONS TO IT
\usepackage{algorithm}
\usepackage{algorithmic}
\usepackage{subfigure}
\usepackage{amsmath}
\usepackage{amsfonts}
\usepackage{color}
\usepackage{booktabs}
\usepackage{makecell}
\usepackage{newfloat}
\usepackage{listings}
\DeclareCaptionStyle{ruled}{labelfont=normalfont,labelsep=colon,strut=off} % DO NOT CHANGE THIS
\floatstyle{ruled}
\newfloat{listing}{tb}{lst}{}
\floatname{listing}{Listing}
\newcommand{\tref}[1]{Table~\ref{#1}}
\newcommand{\eref}[1]{Eq.\eqref{#1}}
\newcommand{\fref}[1]{Fig.~\ref{#1}}

\title{SCP: Spherical-Coordinate-Based Learned Point Cloud Compression}
\author{
    Ao Luo\textsuperscript{\rm 1,2},
    Linxin Song\textsuperscript{\rm 2},
    Keisuke Nonaka\textsuperscript{\rm 1},
    Kyohei Unno\textsuperscript{\rm 1},\\
    Heming Sun\textsuperscript{\rm 3},
    Masayuki Goto\textsuperscript{\rm 2},
    Jiro Katto\textsuperscript{\rm 2}
}
\affiliations{
    \textsuperscript{\rm 1}KDDI Research, Inc.\\
    \textsuperscript{\rm 2}Waseda University\\
    \textsuperscript{\rm 3}Yokohama National University\\
    xao-luo@kddi.com, songlx.imse.gt@ruri.waseda.jp,\\
    ki-nonaka@kddi.com, ky-unno@kddi.com, \\
    sun-heming-vg@ynu.ac.jp, masagoto@waseda.jp, katto@waseda.jp
}

\begin{document}

\maketitle

\begin{abstract}
In recent years, the task of learned point cloud compression has gained prominence. An important type of point cloud, the spinning LiDAR point cloud, is generated by spinning LiDAR on vehicles. This process results in numerous circular shapes and azimuthal angle invariance features within the point clouds. However, these two features have been largely overlooked by previous methodologies.
In this paper, we introduce a model-agnostic method called Spherical-Coordinate-based learned Point cloud compression (SCP), designed to leverage the aforementioned features fully. Additionally, we propose a multi-level Octree for SCP to mitigate the reconstruction error for distant areas within the Spherical-coordinate-based Octree. SCP exhibits excellent universality, making it applicable to various learned point cloud compression techniques. Experimental results demonstrate that SCP surpasses previous state-of-the-art methods by up to 29.14\% in point-to-point PSNR BD-Rate.
% \footnote{The code is available at 
% (after acceptance).
% \url{https://github.com/luoao-kddi/SCP}.
% }
\end{abstract}
\vspace{-3mm}

\section{Introduction}
\label{sec:intro}

LiDAR point clouds play a crucial role in numerous real-world applications, including self-driving vehicles~\cite{selfdriving3, selfdriving, selfdriving2}, robotics~\cite{robot, robot2, robot3} and 3D mapping~\cite{scene, scene2}. However, the transmission and storage of these point clouds present significant challenges. A typical large point cloud may contain up to a million points~\cite{icip2019}, making direct storage highly inefficient.

\begin{figure*}[h!]
    \centering
    \subfigure[The Octree structure produced in Cartesian coordinates.]{
    \includegraphics[width=0.195\textwidth]{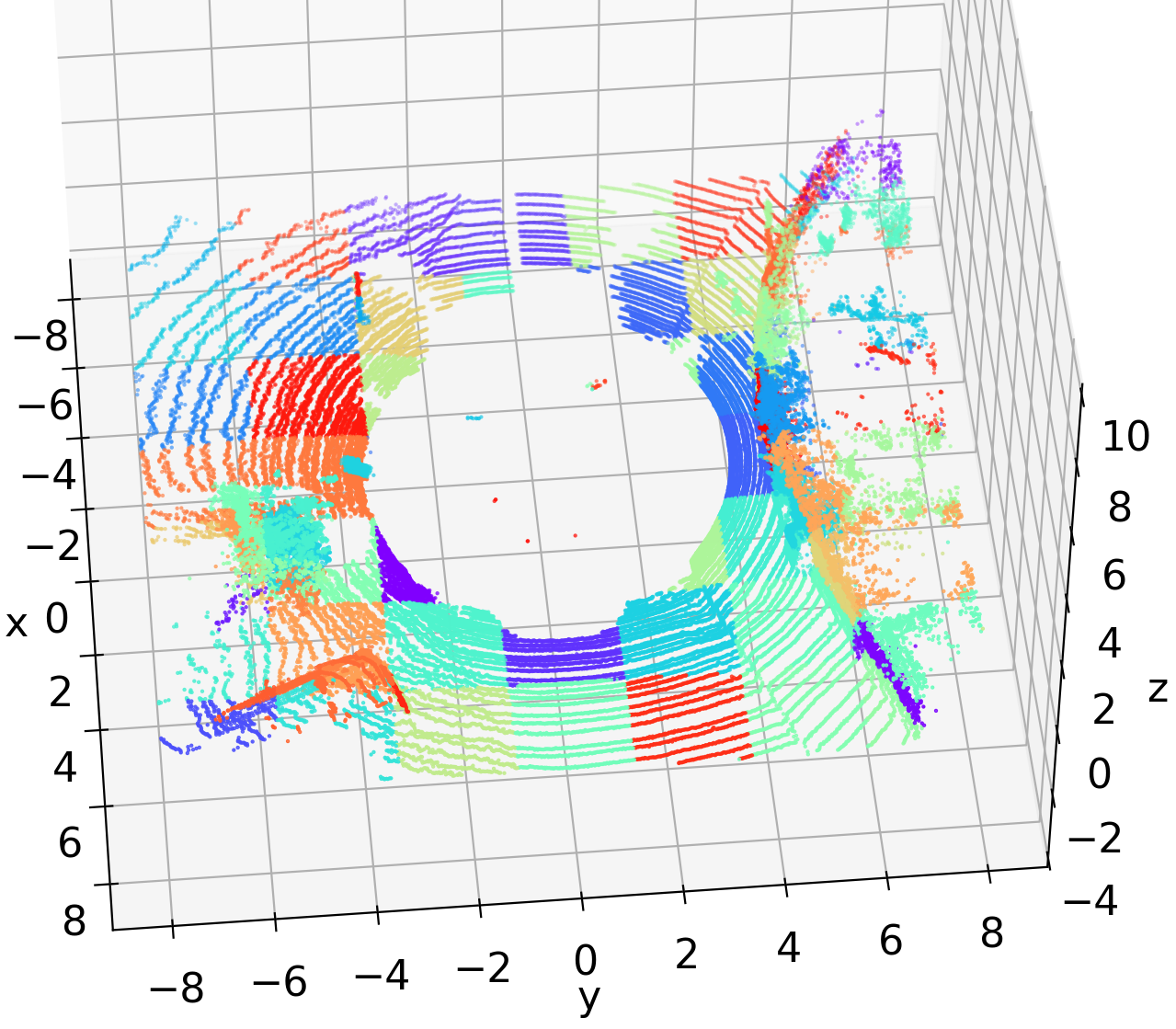}
    \label{fig:cart_split}
    }
    \subfigure[The Octree structure produced in Cylindrical coordinates.]{
    \includegraphics[width=0.45\textwidth]{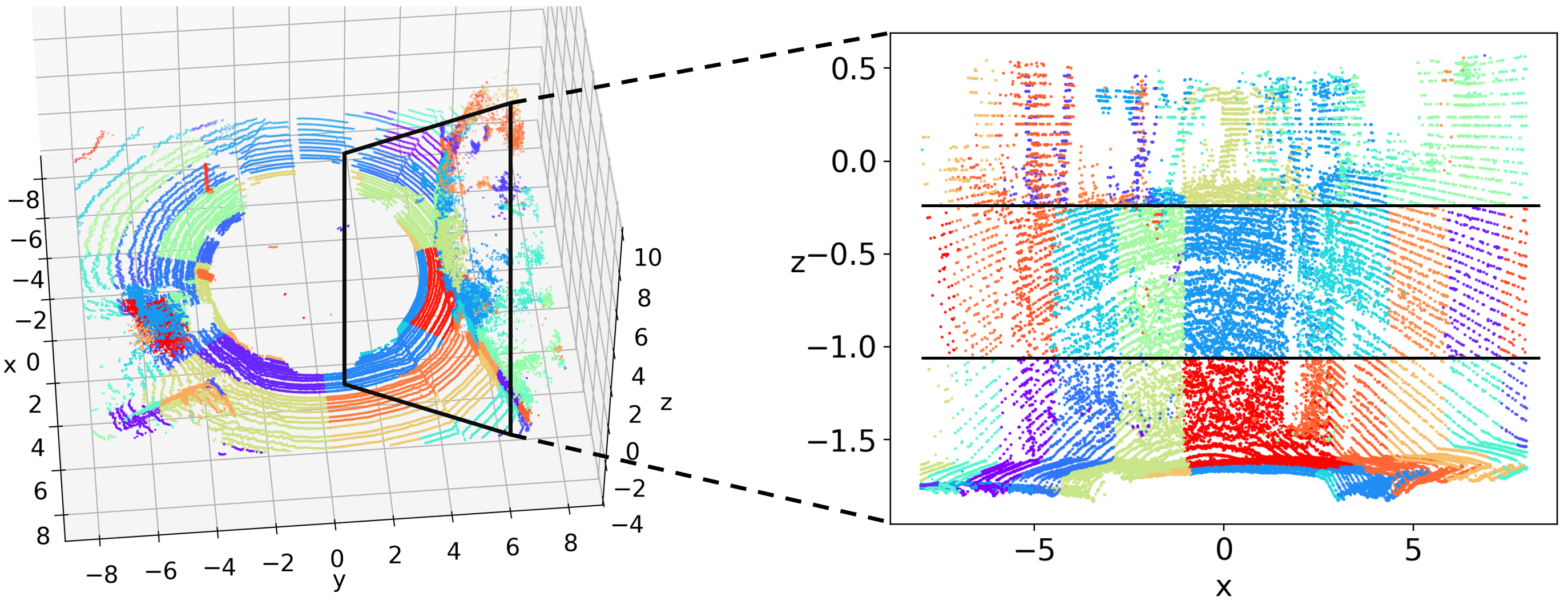}
    \label{fig:cylin_split}
    }
    \subfigure[The Octree structure produced in Spherical coordinates.]{
    \includegraphics[width=0.225\textwidth]{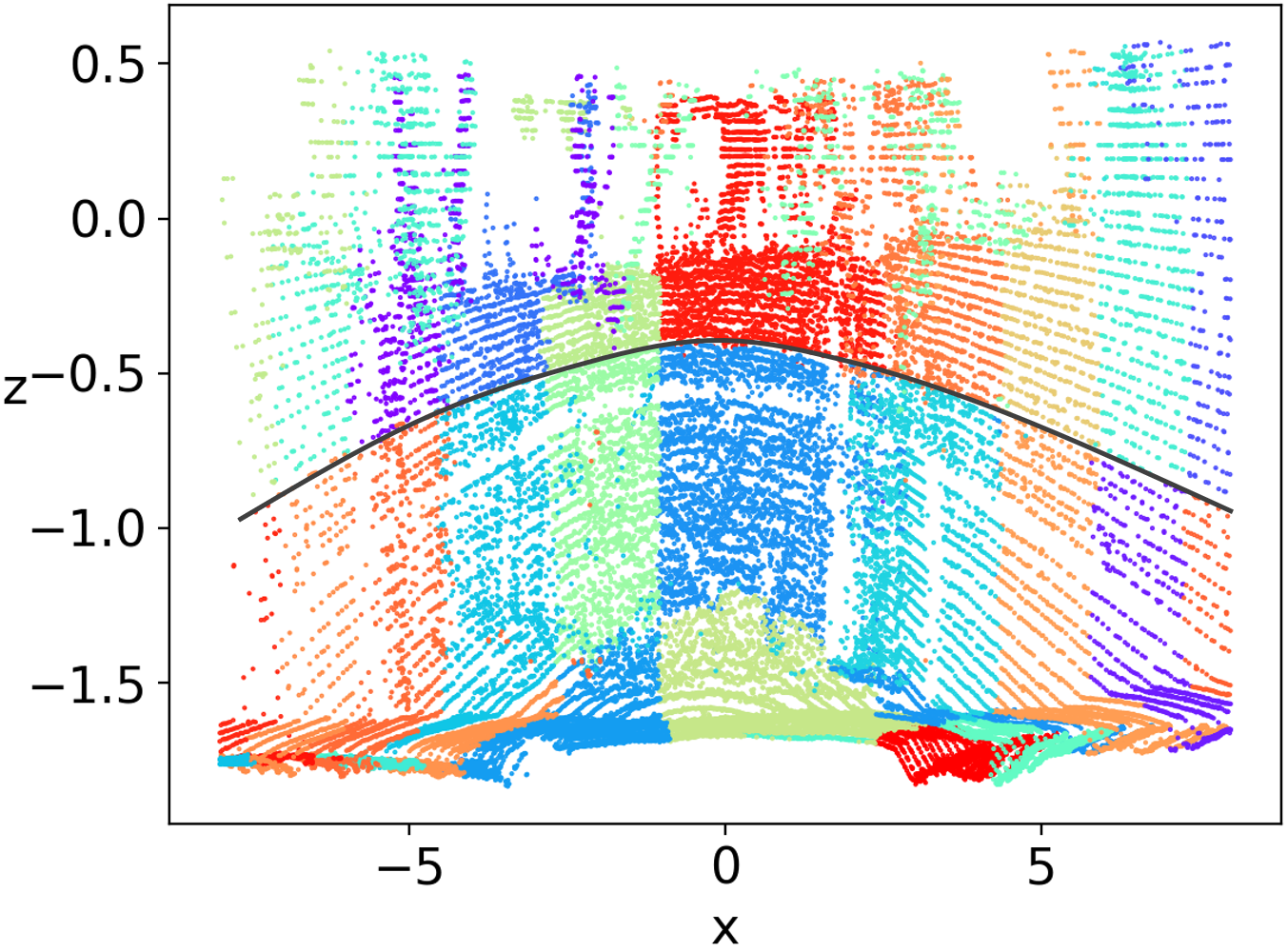}
    \label{fig:spher_split}
    }
    \caption{
    Comparison of the Octree structures in Cartesian, Cylindrical, and Spherical coordinates. The points with the same color are in the same voxel. When looking up-down, the Octree structure in Cylindrical coordinates (see \fref{fig:cylin_split}-left) can better fit the LiDAR point clouds than Cartesian coordinates because it harnesses the circular shapes of LiDAR point clouds. In this up-down view, the Octree structures of Cylindrical and Spherical coordinates are similar. However, when looking from the original point horizontally, as the black lines in \fref{fig:cylin_split}-right and \fref{fig:spher_split}, the Octree in Cylindrical coordinates overlooks the azimuthal angle invariance, often splitting chains into different voxels. In contrast, the Octree in Spherical coordinates tends to group points from the same chain into the same voxel, increasing the relevant information for every point.}
    \label{fig:split}
    \vspace{-5mm}
\end{figure*}

In an effort to mitigate the transmission and storage costs associated with point clouds, MPEG introduced a hand-crafted compression standard known as Geometry-based Point Cloud Compression (G-PCC)~\cite{gpcc}. This standard employs different geometric structures, including the Octree~\cite{octree} and predictive geometry~\cite{pred}, to compress point clouds.
The Octree-based entropy model is a widely adopted approach in both G-PCC and learned methods for representing and compressing point clouds.
Simultaneously, another geometric structure, predictive geometry, capitalizes on the chain structure formed by each LiDAR laser beam. It predicts the position of the next point based on the angles and distances of preceding points on the chain.
Other than basic G-PCC, a Cylindrical-coordinate-based method~\cite{ortega} suggests transforming the Cartesian-coordinate positions into Cylindrical coordinates for G-PCC to compress.
In the Cylindrical coordinates, when splitting $\theta$ coordinate for Octree construction,
points from the same chain (points acquired by the same laser beam) tend to be grouped into the same voxel, as shown in \fref{fig:cylin_split}-left. Thus, these points have more relevant information in their context for compressing. Consequently, this method
improves the performance of G-PCC. In the meanwhile, Cylinder3D~\cite{cylinder3d} also took the use of Cylindrical coordinates in the point cloud segmentation task.
Concurrently, learned point cloud compression methods are emerging, using deep learning techniques to compress point clouds. Former work such as OctSqueeze~\cite{octsqueeze}, VoxelDNN~\cite{voxeldnn}, VoxelContext-Net~\cite{voxelcontext}, and OctFormer~\cite{octformer} employ information of ancient voxels for prediction of the current one.
Advancing these approaches, OctAttention~\cite{octattention}, SparsePCGC~\cite{nju}, and EHEM~\cite{ehem} harness the voxels in the same level as the current one to minimize the redundancy.

% Nevertheless, all the aforementioned learned methods overlook a crucial feature of spinning LiDAR point clouds. \fref{fig:cart_split} visualizes a spinning LiDAR point cloud, which is scanned by a spinning LiDAR. This LiDAR results in numerous circular-shaped point chains within the point clouds, leading to high redundancy.
% The predictive geometry in G-PCC utilizes spherical coordinates for encoding point chains, but it solely uses the information of points within the same chain as references, disregarding other neighboring points.
% The Surface Prior~\cite{surface} uses a geometric prior to normalize predicted points to their neighbors' circular shapes via a loss function, but its vague use of circular shapes and limited context size underutilizes the circular feature in spinning LiDAR point clouds.
% The Cylindrical coordinates~\cite{ortega} inadequately represent point clouds since points in the same chain have different $z$ coordinates, causing separation in different voxels. Fortunately, inspired by predictive geometry, we find these points share the same azimuthal angle, leading to the use of Spherical coordinates where the azimuthal angle replaces the $z$ coordinate. This change allows points in the same chain to likely be in the same voxel when constructing Octree, concentrating relevant information. Thus, Spherical coordinates represent spinning LiDAR point clouds more precisely than Cylindrical coordinates.

Nevertheless, the aforementioned learned methods overlook a crucial feature of spinning LiDAR point clouds. \fref{fig:cart_split} visualizes a spinning LiDAR point cloud. This LiDAR generates numerous circular-shaped point chains within the point clouds, leading to high redundancy.
The predictive geometry in G-PCC utilizes spherical coordinates for encoding point chains, but it solely uses the information of points within the same chain as references, disregarding other neighboring points.
% The Surface Prior~\cite{surface} uses a geometric prior to normalize predicted points to their neighbors' circular shapes via a loss function, but its vague use of circular shapes and limited context size underutilizes the circular feature in spinning LiDAR point clouds.
The Cylindrical coordinates~\cite{ortega} inadequately represent point clouds since points in the same chain have different $z$ coordinates, causing separation in different voxels, as shown in \fref{fig:cylin_split}-right.
Fortunately, inspired by predictive geometry, we find these points share the same azimuthal angle, leading to the use of Spherical coordinates where the azimuthal angle replaces the $z$ coordinate.
This transformation allows points in the same chain to be split in the same voxel, concentrating the relevant information, as shown in \fref{fig:spher_split}.

In this paper, we focus on the spinning LiDAR point cloud, which is called LiDAR PC for abbreviation. To fully leverage the circular-shaped azimuthal-angle-invariant point chains in LiDAR PCs, we introduce a model-agnostic Spherical-Coordinate-based learned Point cloud compression (SCP) method.
SCP transforms LiDAR PCs from Cartesian coordinates to Spherical coordinates, concentrating the relevant information and making it easier for neural networks to reduce redundancy.
% circular shapes and the azimuthal angle invariance
% It gathers relevant voxels together more efficiently than both Cartesian and Cylindrical coordinates.
It's important to note that, as SCP primarily alters the data pre-processing methods, it is model-agnostic and can be applied to various learned point cloud compression methods. In our experiments, we demonstrate its effectiveness by applying SCP to recent Cartesian-coordinate-based methods like EHEM~\cite{ehem} and OctAttention~\cite{octattention}.
Additionally, as shown in \fref{fig:spher_sparse}, the distant voxels in the Spherical-coordinate-based Octree have larger voxel sizes than the central voxels, and the voxel size is positively correlated with the reconstruction errors, resulting in higher reconstruction errors for distant voxels.
To solve this problem, we propose a multi-level Octree for SCP, which assigns additional levels to distant voxels. With more levels, the voxel size decreases, leading to lower reconstruction errors.

We trained and evaluated the baseline methods and SCP on the SemanticKITTI~\cite{kitti} and Ford~\cite{ford} datasets. According to our experiment results, SCP surpasses previous state-of-the-art methods in all conducted experiments. The contributions of our work can be summarized as follows:
\begin{itemize}
    \item We introduce a model-agnostic Spherical-Coordinate-based learned Point cloud compression (SCP) method. This method transforms point clouds into Spherical coordinates, thereby fully leveraging the circular shapes and the azimuthal angle invariance feature inherent in LiDAR PCs.
    \item We propose a multi-level Octree structure, which mitigates the increment of reconstruction error of distant voxels in the Spherical-coordinate-based Octree.
    \item Our experiments on various backbone methods demonstrate that our methods can efficiently reduce redundancy among points, surpassing previous state-of-the-art methods by up to 29.14\% in point-to-point PSNR BD-Rate~\cite{bdrate}.
\end{itemize}

\vspace{-2mm}

\section{Related Work}
\label{sec:relate}

\subsection{Hand-Crafted Methods}

Hand-crafted methods based on geometry typically employ tree structures like Octree~\cite{octree, rw1, unno2023} and predictive geometry~\cite{pred} to organize unstructured point clouds. The Octree structure is widely used in numerous methods, such as \cite{rw2, rw3}. Predictive geometry, on the other hand, models LiDAR PCs as a predictive tree, where each branch represents a complete point chain scanned by a laser beam. This approach leverages the circular chain structure in LiDAR PCs to predict subsequent points based on the angles and distances of previous points in the chain.
Leveraging the circular shape of LiDAR PCs, \cite{ortega} proposed a transformation from Cartesian-coordinate-based positions to Cylindrical coordinates. This approach, being sensitive to circular shapes, enhances the performance of the G-PCC algorithm in compressing point clouds.

\subsection{Learned Point Cloud Compression Methods}

In recent years, learned point cloud compression methods have been emerging. Many of these techniques, including those cited in \cite{nguyen2021learning, icip2019, voxelcontext, voxeldnn, nju}, utilize Octree to represent and compress point clouds.

OctSqueeze~\cite{octsqueeze} builds the Octree of the point cloud, predicting voxel occupancy level by level, using information from ancient voxels and known data about the current voxel.
Building upon OctSqueeze, methods such as VoxelDNN~\cite{voxeldnn}, VoxelContext-Net~\cite{voxelcontext}, SparsePCGC~\cite{nju}, and OctFormer~\cite{octformer} eliminate redundancy by employing the information of neighbor voxels of the parent voxel.
%, thereby shaping the context of the current voxel into a cube.
Moreover, Surface Prior~\cite{surface} incorporates neighbor voxels which share the same depth as the current coding voxel, into the framework. OctAttention~\cite{octattention} further utilizes this kind of voxel, increasing the context size from surrounding 26 voxels to 1024 voxels. This significantly expands the receptive field of the current voxel.
Building on OctAttention, EHEM~\cite{ehem} further exploits the potential of the Transformer framework by integrating the Swin Transformer~\cite{swin} into their method. This effectively increases the context size to 8192 sibling voxels. Simultaneously, it transitions from serial coding to a checkerboard~\cite{checkerboard} type, resulting in a substantial reduction in decoding time.
In another aspect, \cite{surface} employs a context with uncle voxels and utilizes the feature of circular shapes in the LiDAR PCs by incorporating a geometry-prior loss for training.

\section{Preliminary}
\label{sec:pre}

In this section, we explain the Octree structure, which efficiently represents the point clouds. Then the coordinate systems involved in our proposed method are introduced. Finally, our optimizing target is illustrated.

\subsection{Octree Structure}

The Octree is a hierarchical data structure that provides an efficient way of representing 3D point clouds and is beneficial for dealing with inherently unstructured and sparse LiDAR PCs.
% The strength of Octree lies in its progressive structure. Every level of the Octree symbolizes the point cloud with a different resolution.
When constructing the Octree, the whole point cloud is taken as a cube, a.k.a a voxel, then divided into 8 equal-sized sub-voxels. This procedure repeats until the side length of a leaf voxel is equal to the quantization step. Then the quantization operation merges points in the same leaf voxel together. In each non-leaf voxel, the sub-voxels with points in them are represented by a bit $1$, and the empty ones are set to value $0$. Therefore, each non-leaf voxel is represented by an occupancy symbol composed of 8 bits $(1-255)$, where each bit indicates the occupancy status of the corresponding sub-voxel.

\subsection{Coordinate Systems}

We introduce different coordinate systems involved in our experiments: 3-D Cartesian coordinates, Cylindrical coordinates, and Spherical coordinates.
% The Cartesian coordinates are oriented along straight lines, while cylindrical and spherical coordinates are oriented around a point. This makes Cartesian coordinates more suitable for problems involving linear motion, while cylindrical and spherical coordinates are more suitable for problems involving circular or rotational motion.

\subsubsection{3-D Cartesian Coordinate System} describes each point in the space by 3 numerical coordinates, which are the signed distances from the point to three mutually perpendicular planes. These coordinates are given as $(x, y, z)$.

\subsubsection{Cylindrical Coordinate System} is a natural extension of polar coordinates. It describes a point by three values: radial distance $\rho$ from the origin, the angular coordinate $\theta$ (the same as in polar coordinates), and the height $z$, which is the same as in Cartesian coordinates. The transformation from Cartesian coordinates to Cylindrical coordinates can be written as:

\begin{equation}
\begin{aligned}
    \rho = \sqrt{x^2+y^2},
    \theta = \arctan\frac{y}{x},
    z = z,
\end{aligned}
\end{equation}
note that the quadrant of $\theta$ is decided by signs of $x$ and $y$.

\begin{figure}[h!]
    \centering
    \includegraphics[width=0.4\textwidth]{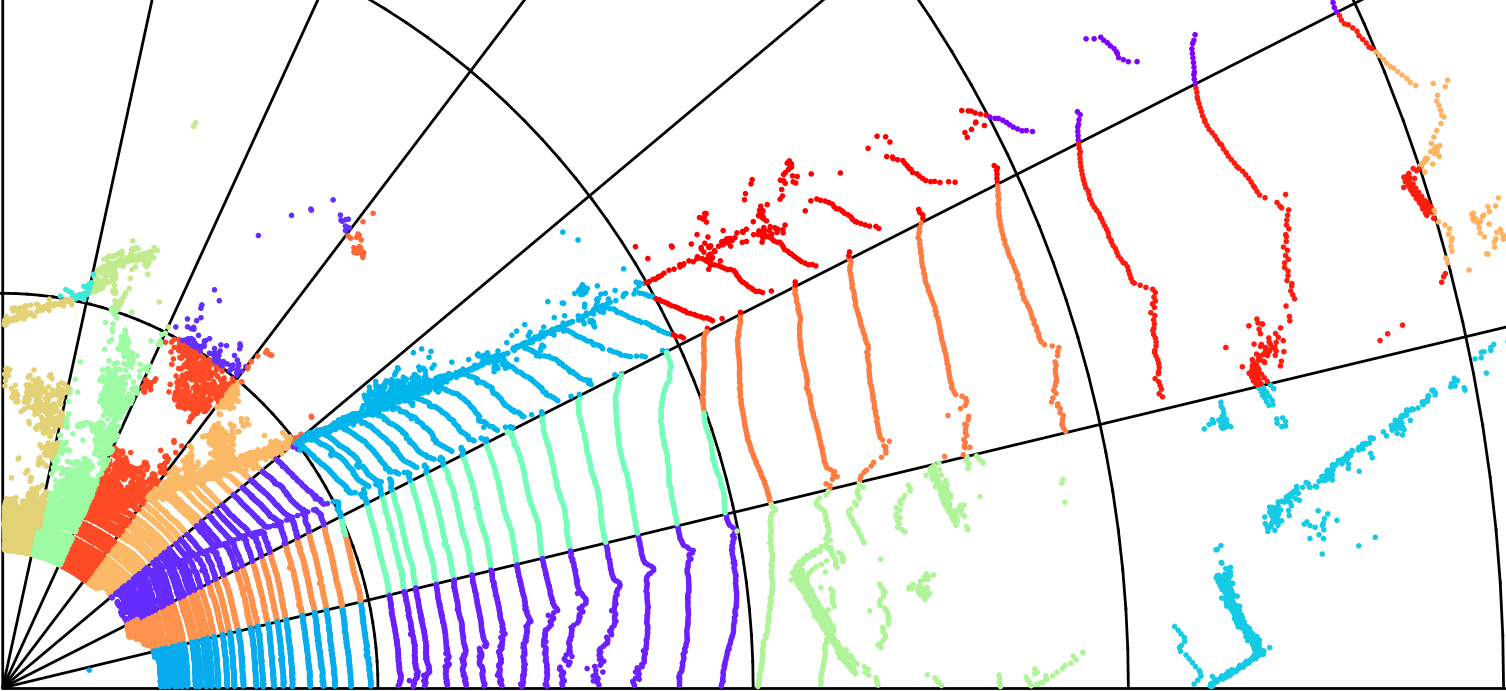}
    \caption{
    In Spherical coordinates, the size of the voxels varies based on their distance from the origin (left-bottom corner). We distinguish voxels by different colors. The sizes of distant voxels are larger than those of the central ones. This phenomenon results in lower reconstruction errors for the central voxels and higher errors for the distant ones.}
    \label{fig:spher_sparse}
\end{figure}

\subsubsection{Spherical Coordinate System} also describes a point with three values: the radial distance $\rho$ from the origin to the point, and two angles: the polar angle $\theta$, which is the same angle used in cylindrical coordinates, and the azimuthal angle $\varphi$, which is the angle from the positive $z$ coordinate to the point. The transformation equations from Cartesian coordinates are:

\begin{equation}
\begin{aligned}
    \rho = \sqrt{x^2+y^2+z^2},
    \theta = \arctan\frac{y}{x},
    \varphi = \arccos\frac{z}{\rho},
\end{aligned}
\end{equation}
again, the quadrant of $\theta$ depends on the signs of $x$ and $y$.

\subsection{Task Definition}

The occupancy sequence $\boldsymbol{x}=\{x_1, ..., x_V\}$ ($V$ is the number of voxels) of voxels in an Octree is losslessly encoded by an entropy encoder into a bitstream, based on the predicted probability from a neural network. After transmission, the Octree is reconstructed from the entropy-decoded occupancy sequence and converted back into the point cloud.
Hence, the objective of point cloud geometry compression is to predict the occupancy of each voxel in the Octree, which can be viewed as a 255-category classification problem. Given all the parameters in our model as $\boldsymbol{w}$ and context information as $\boldsymbol{C}=\{\boldsymbol{C}_1, ..., \boldsymbol{C}_V\}$, the optimization objective $\mathcal{L}$ is the cross-entropy between the ground truth occupancy value $t\in[1, 255]$ and the corresponding predicted probability of each voxel $\tilde{p}_t(x_i|\boldsymbol{C}_i;\boldsymbol{w})$. This can be defined as follows:

\begin{equation}
\begin{aligned}
    \mathcal L = -\sum_i^V\log \tilde{p}_t(x_i|\boldsymbol{C}_i;\boldsymbol{w}).
\end{aligned}
\end{equation}

\section{Methodology}
\label{sec:method}

We propose a model-agnostic Spherical-Coordinate-based learned Point cloud compression (SCP) method, which efficiently leverages the features of LiDAR PCs.
Furthermore, we introduce a multi-level Octree method to solve the higher reconstruction errors problem for distant voxels in the Spherical-coordinate-based Octree.

\subsection{Spherical-Coordinate-based Learned Point Cloud Compression}

The circular shapes and azimuthal angle invariance features in LiDAR PCs lead to a high degree of redundancy.
In common-used Cartesian coordinates, redundant points are segmented into discrete voxels, as shown in \fref{fig:cart_split}. This segmentation creates an unstable context, in which nearby voxels may not be relevant.
To leverage both features of LiDAR PCs, we propose the Spherical-Coordinate-based learned Point cloud Compression (SCP) method, which converts the points from Cartesian coordinates to Spherical coordinates.
% leveraging the features of circular shapes and azimuthal angle invariance to consolidate relevant voxels in the same chain together.
When looking up-down, the Spherical-coordinate-based Octree has the same structure as the Cylindrical one, as shown in \fref{fig:cart_split}-left. Points in the same chain are assigned together, making better usage of circular shapes.
On the other hand, when looking horizontally from the original point, the Spherical-coordinate-based Octree (\fref{fig:spher_split}) is more likely to assign the points in the same chain to the same voxel compared with the Cylindrical one (\fref{fig:cart_split}-right), harnessing the azimuthal angle invariance feature.
Furthermore, the representation of circular structures in Spherical coordinates is simplified from quadratic to linear, making the prediction problem easier to solve, as illustrated in \fref{fig:linear}.

When constructing Octree in Spherical coordinates, the quantization procedure differs from that in Cartesian coordinates. The variation range for the $\rho$, $\theta$ and $\varphi$ coordinates in Spherical coordinates are $[0, \rho_{max}]$, $[0, 2\pi)$ and $[0, \pi]$, respectively. The $\rho_{max}$ varies in different point clouds. When quantizing the point clouds, we adopt the quantization step $q$ of EHEM~\cite{ehem} to the $\rho$ coordinate, noted as $q_\rho=q$. Therefore, the total number of bins for $\rho$ coordinate is $b=\lceil\rho_{max} / q_{\rho}\rceil$. Subsequently, we divide the ranges of the $\theta$ and $\varphi$ coordinates by $b$ separately to generate their respective quantization steps, as demonstrated in \eref{eq:qs}.

\begin{equation}
\begin{aligned}
    q_{\theta} &= 2\pi / (b - 1)\approx\frac{2\pi q}{\rho_{max}},\\
    q_{\varphi} &= \pi / (b - 1)\approx\frac{\pi q}{\rho_{max}},\\
    \label{eq:qs}
\end{aligned}
\vspace{-2mm}
\end{equation}
where $q$ is the quantization step. By setting these $q_\theta$ and $q_\varphi$, the bin number of each coordinate are the same.

% ----------------------------------------------
\subsection{Multi-Level Octree for SCP}\label{sec:mullevel}

In the Spherical-coordinate-based Octree, the distance $\rho$ is positively related to the reconstruction error, causing the higher reconstruction error of distant voxels. To maintain a reconstruction error lower or similar to that in Cartesian coordinates, we propose a multi-level Octree for SCP, which allocates additional levels to distant voxels. We initially discuss the relation between distant voxels and reconstruction errors of the quantized point cloud in Cartesian and Spherical coordinates, then illustrate our multi-level Octree method.

For Cartesian coordinates, the upper bound of the error can be expressed as follows:

\begin{figure}[h!]
    \centering
    \includegraphics[width=0.3\textwidth]{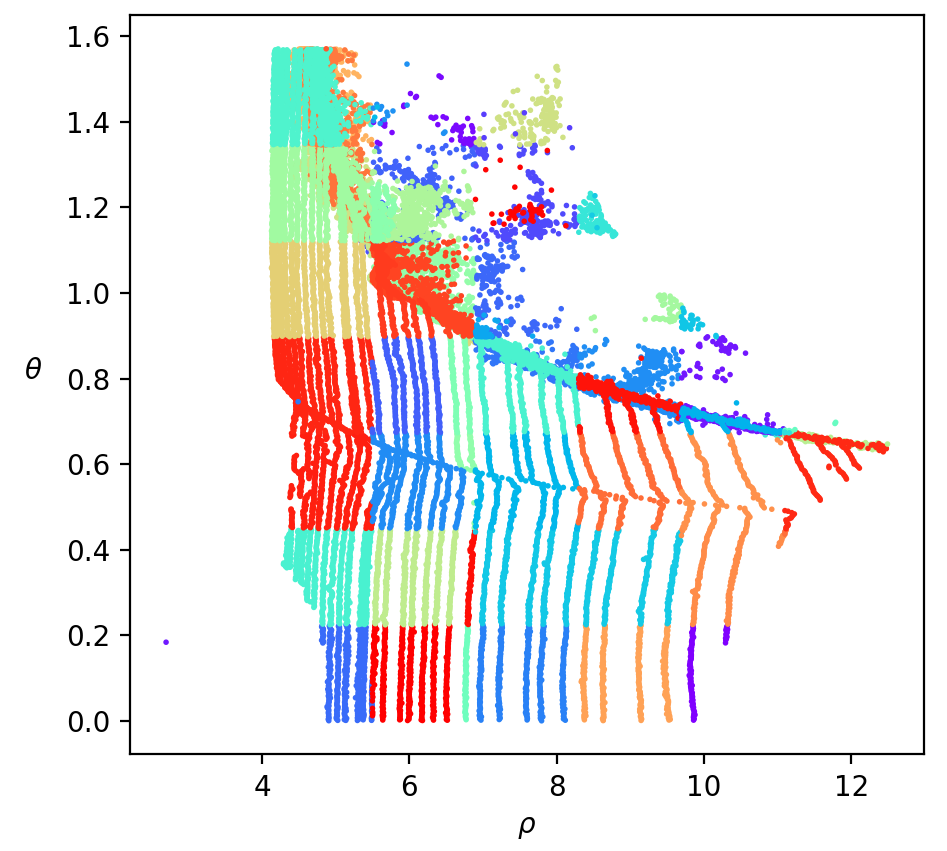}
    \caption{
    % Projections on $\rho o\theta$-plane of point positions in Spherical coordinates. The left parts, which are circular shapes in the point cloud, are nearly converted to straight lines and are easier to be utilized.
    The projections of point positions in Spherical coordinates onto the $\rho o\theta$-plane. The left parts, which represent circular shapes in the point cloud, are nearly straight lines, which is simpler for the compression of points.}
    \label{fig:linear}
    \vspace{-6mm}
\end{figure}

\begin{equation}
\begin{aligned}
    \varepsilon_c = \max_i\|p_i - \hat p_i\|_{2} \leq \frac{\sqrt{3}q}{2},
    \label{eq:error_c}
\end{aligned}
\end{equation}
where $p_i$ represents the original position of point $i$, and $\hat p_i$ is its position after quantization, $q$ is the quantization step size. It is obvious that this error is uniformly distributed throughout the entire space of Cartesian coordinates because the upper bound is only related to the quantization step. The complete derivation can be found in the Appendix.

In the Spherical coordinate system, however, the voxels' reconstruction error is non-uniform.
% and will increase together with the distance from the center.
% To demonstrate this, we first assume that we have two points with fixed $\theta$ and $\varphi$ values: $p_1 = (\rho, \theta_1, \varphi_1)$ and $p_2 = (\rho, \theta_2, \varphi_2)$. It is easy to derive that the Euclidean distance between $p_1$ and $p_2$ positively relates to the value of $\rho$. Therefore, when transforming the point clouds from Cartesian to Spherical coordinates, the sizes of the voxels in the Octree become various.
As depicted in \fref{fig:spher_sparse}, the further a voxel (with a larger $\rho$ value) is from the origin, the larger its size. This characteristic leads to a higher upper bound of reconstruction error for distant voxels. As shown in \eref{eq:error_s}, the reconstruction error $\varepsilon_s$ linearly correlated with the value of the $\rho$ coordinate.

\begin{equation}
\begin{aligned}
    \varepsilon_s = \max_i\|p_i - \hat p_i\|_{2} \leq \frac{\sqrt{5}\pi q}{2\rho_{max}}\rho.
    \label{eq:error_s}
\end{aligned}
\end{equation}
We put the proof into the Appendix and further discuss the quantization results in the Experimental Results of the Experiments section.

\begin{figure*}[h!]
    \centering
    \subfigure[PSNR D1 results on SemanticKITTI.]{
    \includegraphics[width=0.32\textwidth]{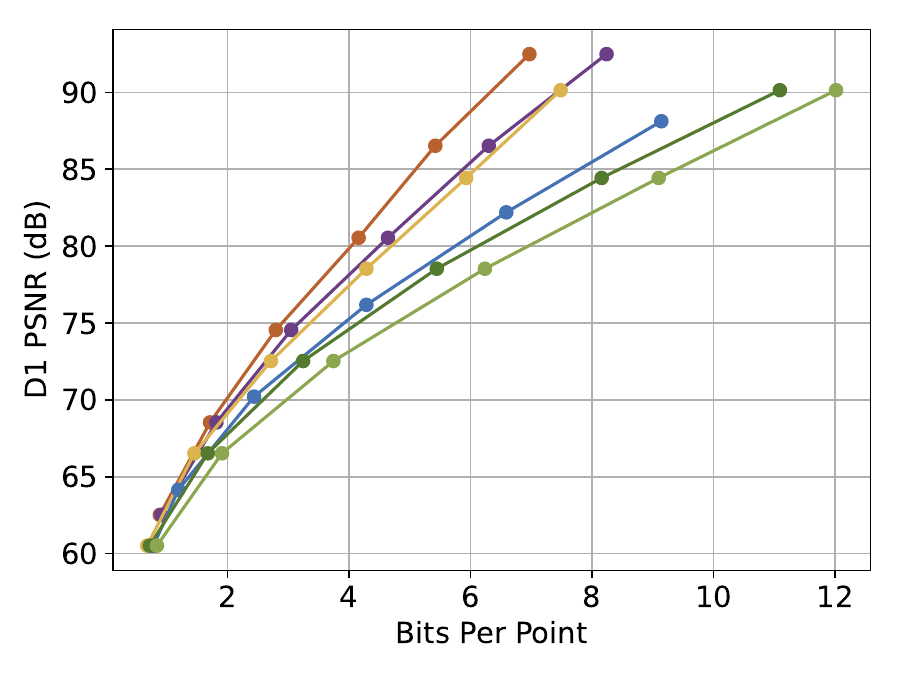}
    }
    \subfigure[PSNR D2 results on SemanticKITTI.]{
    \includegraphics[width=0.32\textwidth]{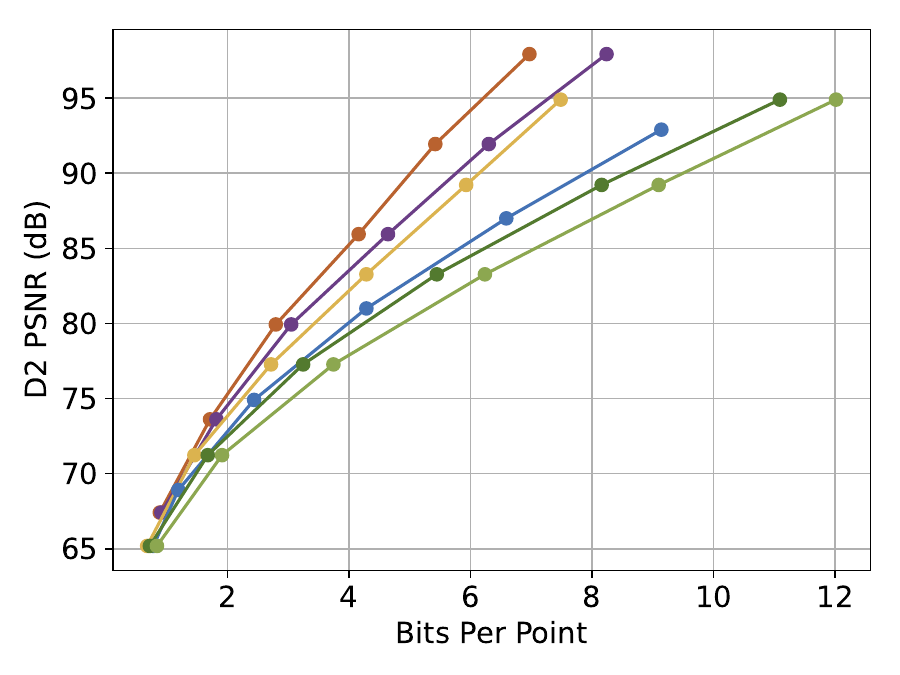}
    }
    \subfigure[CD results on SemanticKITTI.]{
    \includegraphics[width=0.32\textwidth]{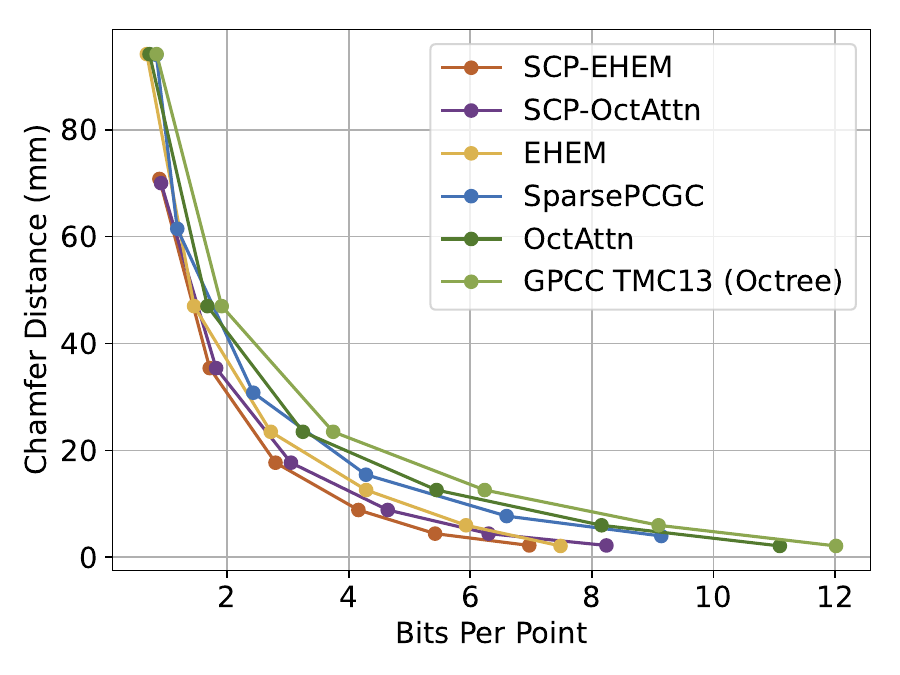}
    }

    \subfigure[PSNR D1 results on Ford.]{
    \includegraphics[width=0.32\textwidth]{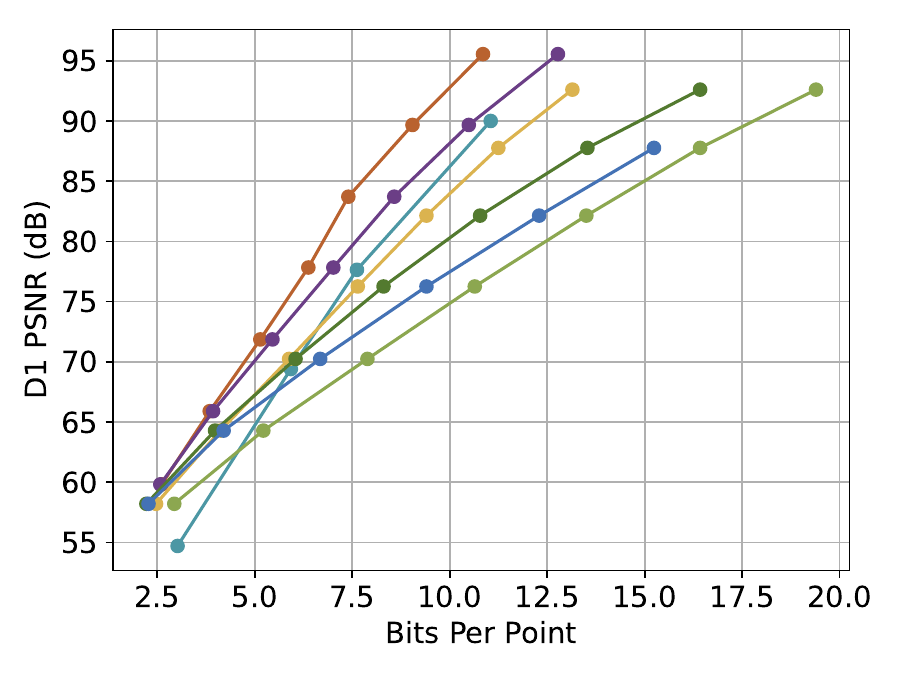}
    }
    \subfigure[PSNR D2 results on Ford.]{
    \includegraphics[width=0.32\textwidth]{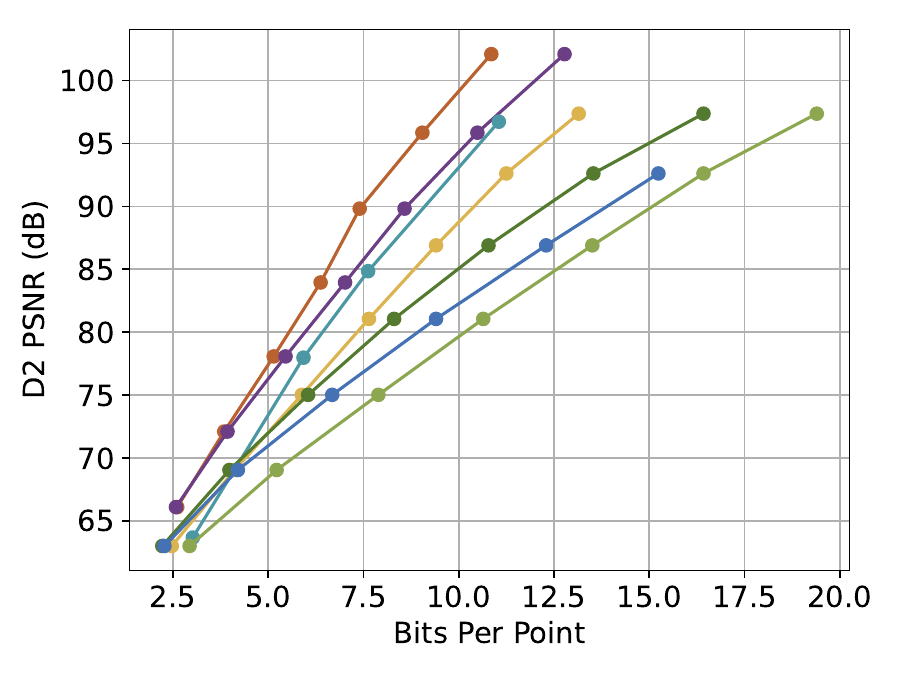}
    }
    \subfigure[CD results on Ford.]{
    \includegraphics[width=0.32\textwidth]{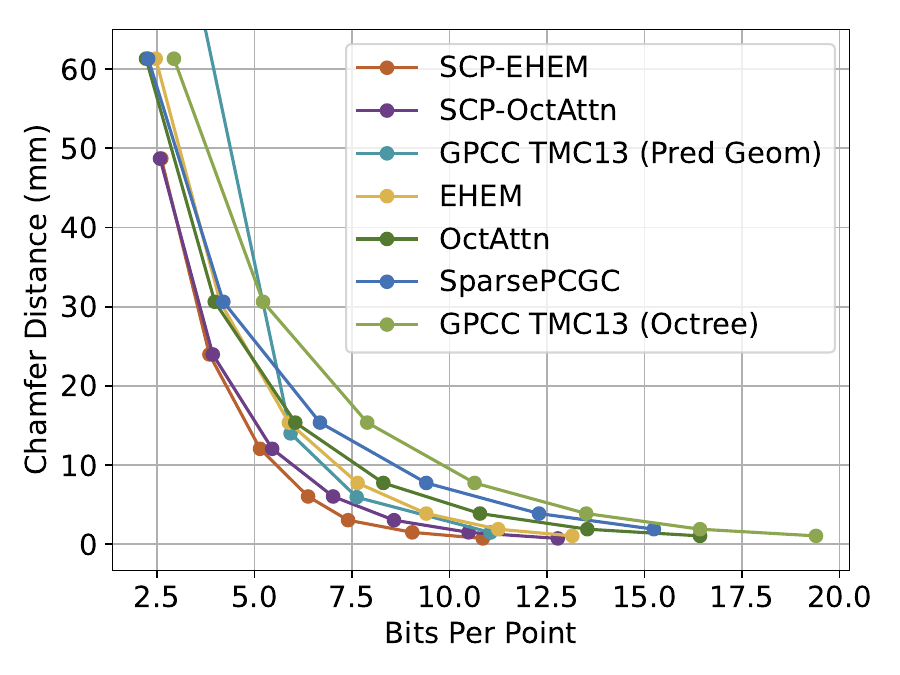}
    }
    \caption{Quantitative rate-distortion results of our SCP-EHEM and SCP-OctAttn models on SemanticKITTI (top) and Ford (bottom) datasets. The baselines are EHEM~\cite{ehem}, SparsePCGC~\cite{nju}, OctAttention~\cite{octattention} and G-PCC TMC13~\cite{gpcc} with either Octree or predictive geometry. Predictive geometry is only available for the Ford dataset, as the SemanticKITTI dataset lacks the necessary sensor information for the calculation of predictive geometry.}
    \label{fig:res}
    \vspace{-4mm}
\end{figure*}

Based on the above findings, if we allocate additional levels (depths) to the distant areas in the Octree, each added level will reduce the quantization step size of the distant area from $q$ to $\frac{1}{2}q$, which in turn halves $\varepsilon_s$ according to \eref{eq:error_s}.
Motivated by this, we propose a multi-level Octree method to reduce the reconstruction error of distant voxels by adding additional levels. Specifically, we divide the point cloud into $N$ parts to assign different numbers of additional levels. Each part has $n\in\{0, 1, ..., N-1\}$ additional levels with $\rho_n\in[t_n\rho_{max}, t_{n+1}\rho_{max})$, where $0 \leq t_n < t_{n+1} < 1$, $t_0 = 0$, $t_N = 1$.
The assigned extra levels can be represented by the reduction of the quantization $q_n = \frac{q}{2^n}$, making the reconstruction error upper bound of each part $n$ lower or similar to the upper bound of $\varepsilon_c$, which can be expressed as

\begin{equation}
\begin{aligned}
    \varepsilon_n \leq \frac{\sqrt{5}\pi q_n}{2\rho_{max}}\cdot\frac{t_n+t_{n+1}}{2}\rho_{max} = \frac{\sqrt{5}\pi q(t_n+t_{n+1})}{2^{n+2}},
    \label{eq:error_n}
\end{aligned}
\end{equation}
where we assign $n$ additional levels for the part $n$. In our experiments, we describe how to pick the optimal $t_n$ for each part to keep $\varepsilon_n\leq\varepsilon_c$.

\section{Experiments}
\label{sec:exp}

\subsection{Experiment Settings}

\subsubsection{Datasets}
We compare all the baseline models and SCP on two LiDAR PC datasets, SemanticKITTI~\cite{kitti} and Ford~\cite{ford}. SemanticKITTI comprises 43,552 LiDAR scans obtained from 22 point cloud sequences. The default split for training includes sequences 00 to 10, while sequences 11 to 21 are used for evaluation. We quantize them with a quantization step of $\frac{400}{2^D-1}$ with Octree depth $D$.
The Ford dataset, utilized in the MPEG point cloud compression standardization, includes three sequences, each containing 1,500 scans. We adhere to the partitioning of the MPEG standardization, where sequence 01 is used for training and sequences 02 and 03 for evaluation. Each sequence in the Ford dataset contains an average of 100,000 points per frame and is quantized to a precision of 1mm. We set the quantization step to $2^{18-D}$ with Octree depth $D$. Both settings are the same as EHEM.

\subsubsection{Metrics}

\begin{figure*}[h!]
    \centering
    \subfigure[EHEM, 90.597dB@7.034.]{
    \includegraphics[width=0.32\textwidth]{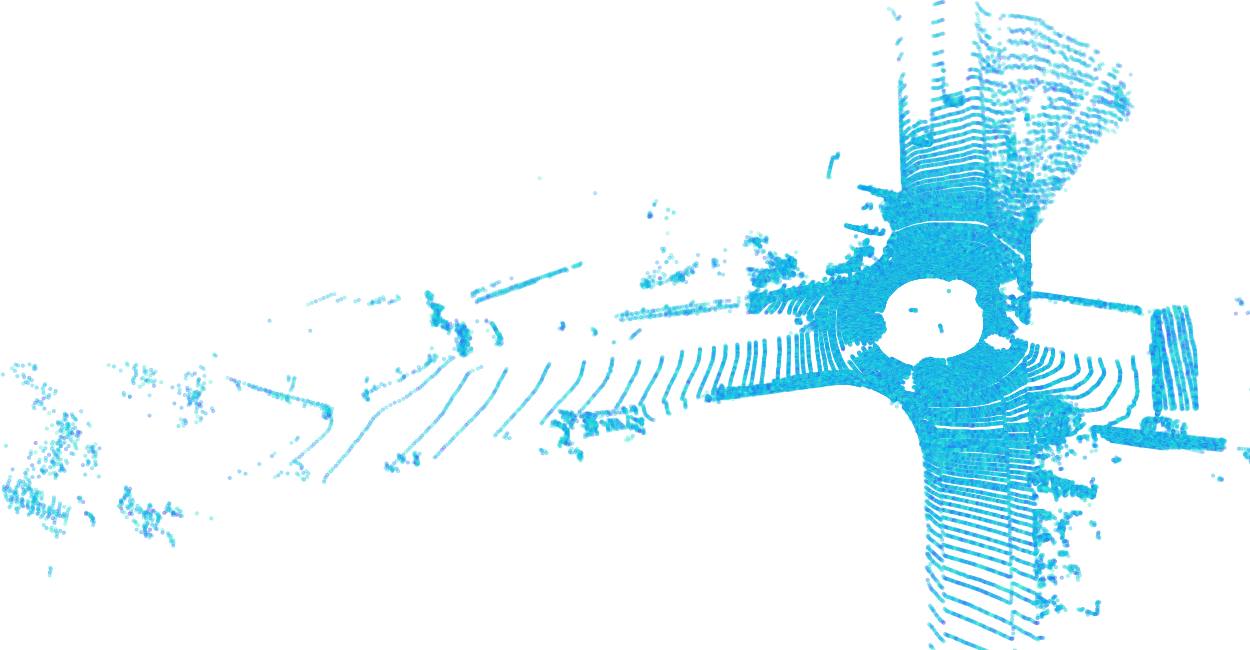}
    \label{fig:vis_cart}
    }
    \subfigure[SCP, 92.624dB@6.069.]{
    \includegraphics[width=0.32\textwidth]{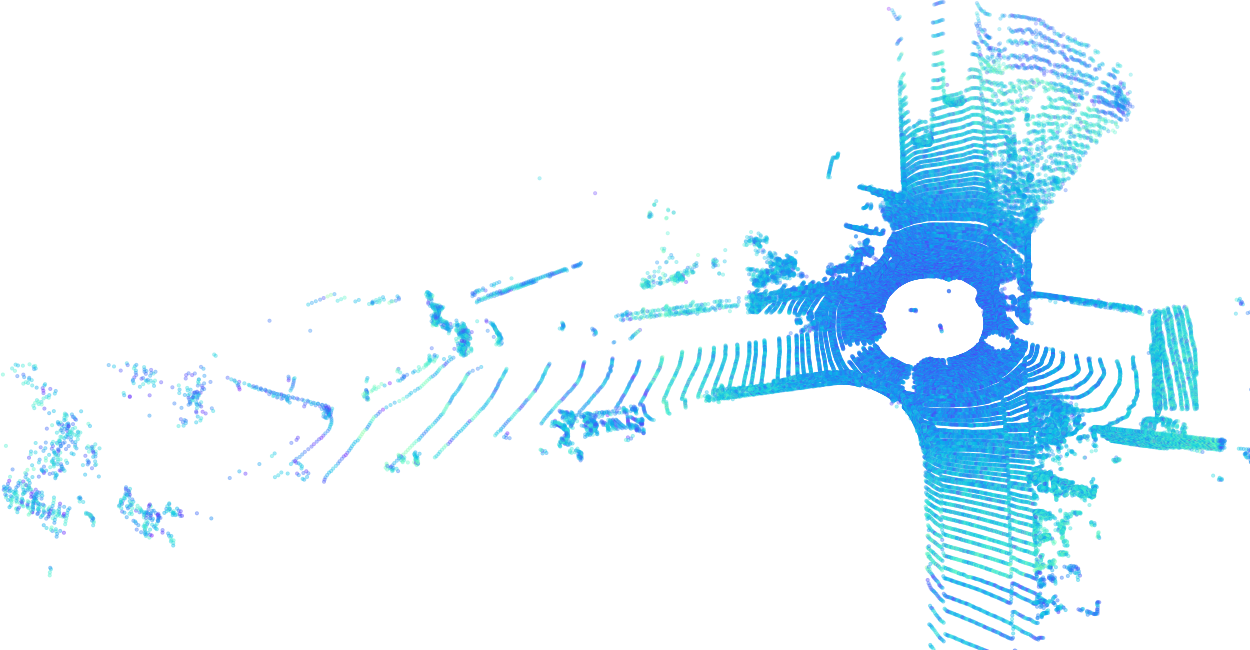}
    \label{fig:vis_mul}
    }
    \subfigure[SCP w/o multi-level, 91.6454dB@5.953.]{
    \includegraphics[width=0.32\textwidth]{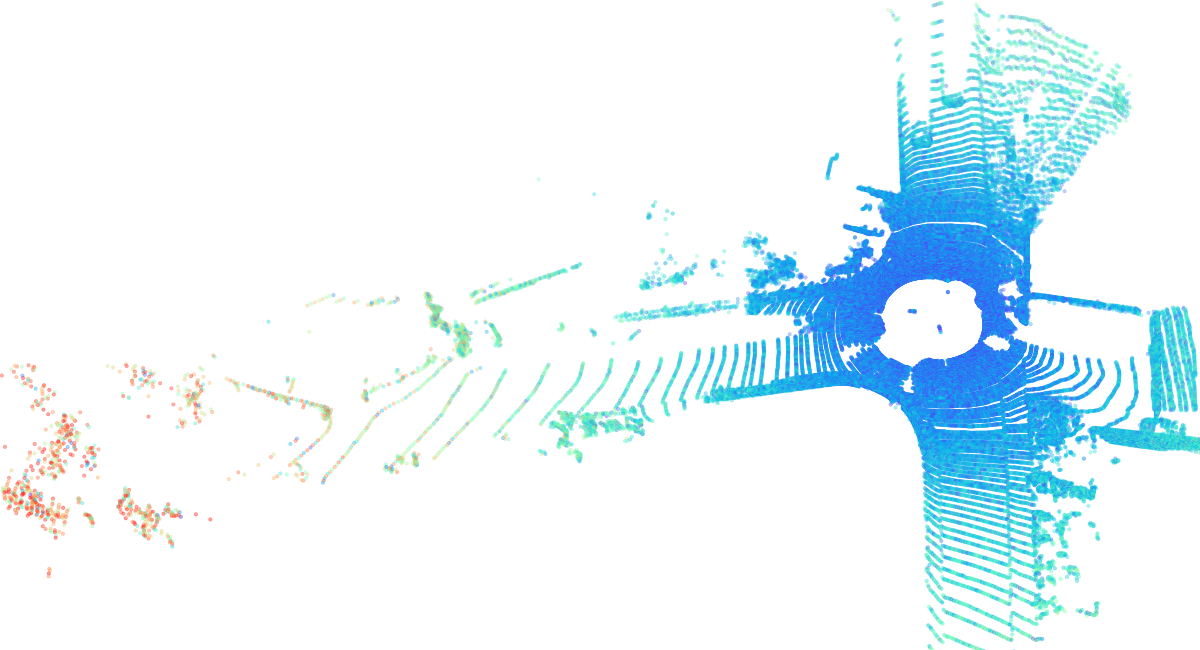}
    \label{fig:vis_same}
    }
    
    \subfigure[EHEM, 90.605dB@7.251.]{
    \includegraphics[width=0.32\textwidth]{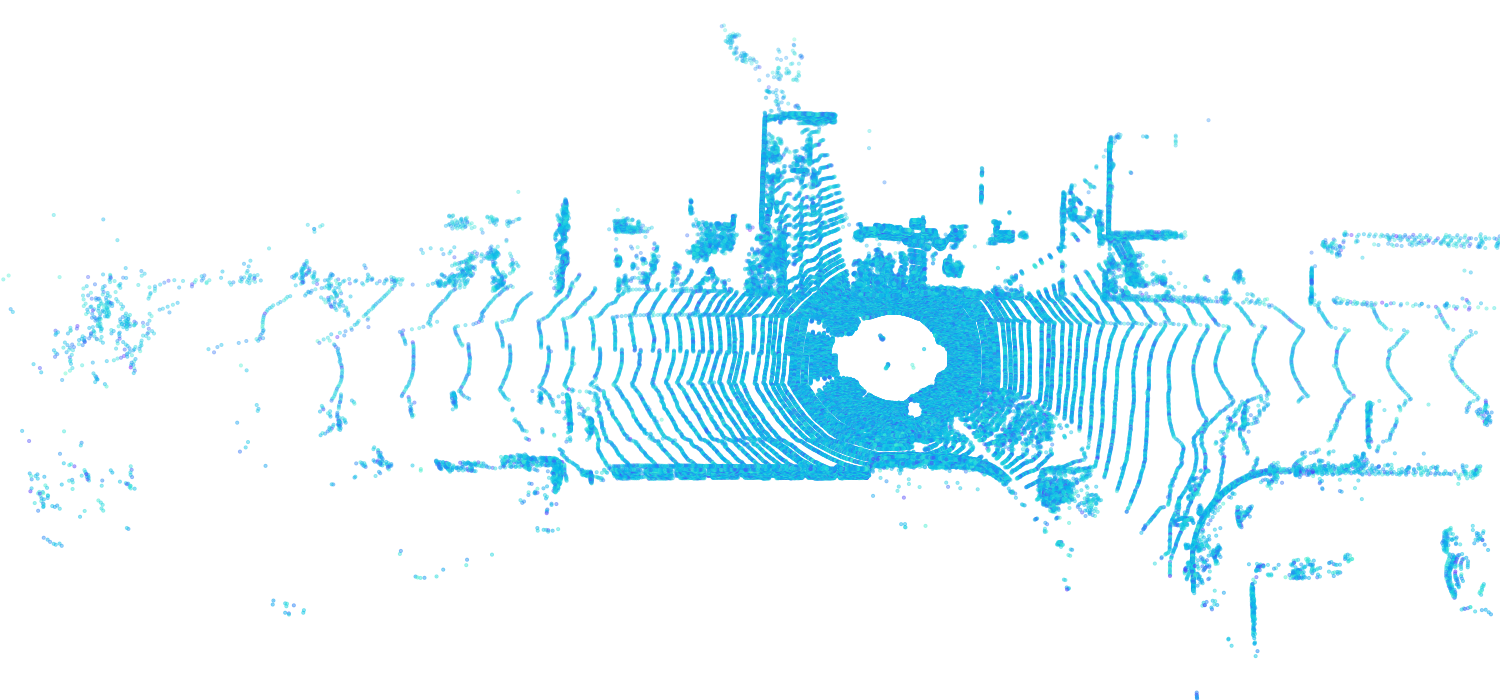}
    \label{fig:vis2_cart}
    }
    \subfigure[SCP, 92.126dB@6.324.]{
    \includegraphics[width=0.32\textwidth]{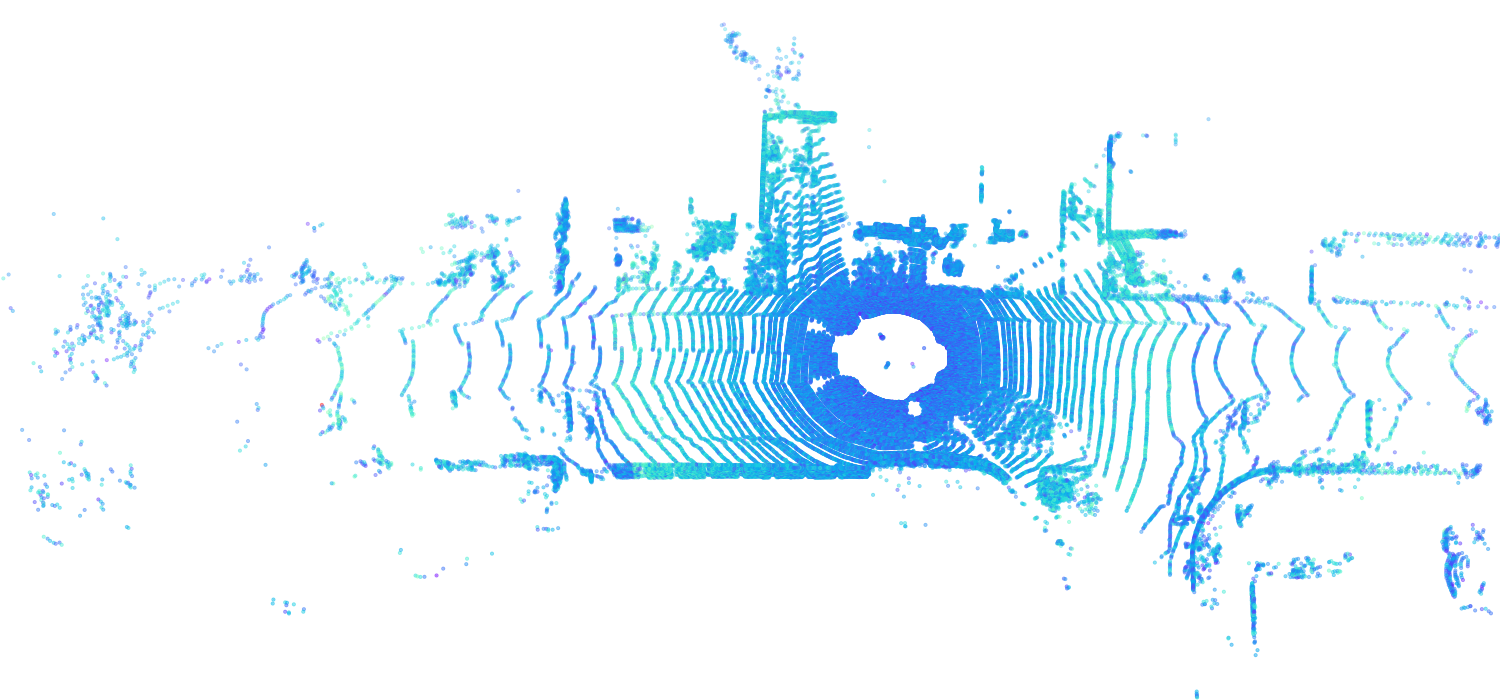}
    \label{fig:vis2_mul}
    }
    \subfigure[SCP w/o multi-level, 90.771dB@6.104.]{
    \includegraphics[width=0.32\textwidth]{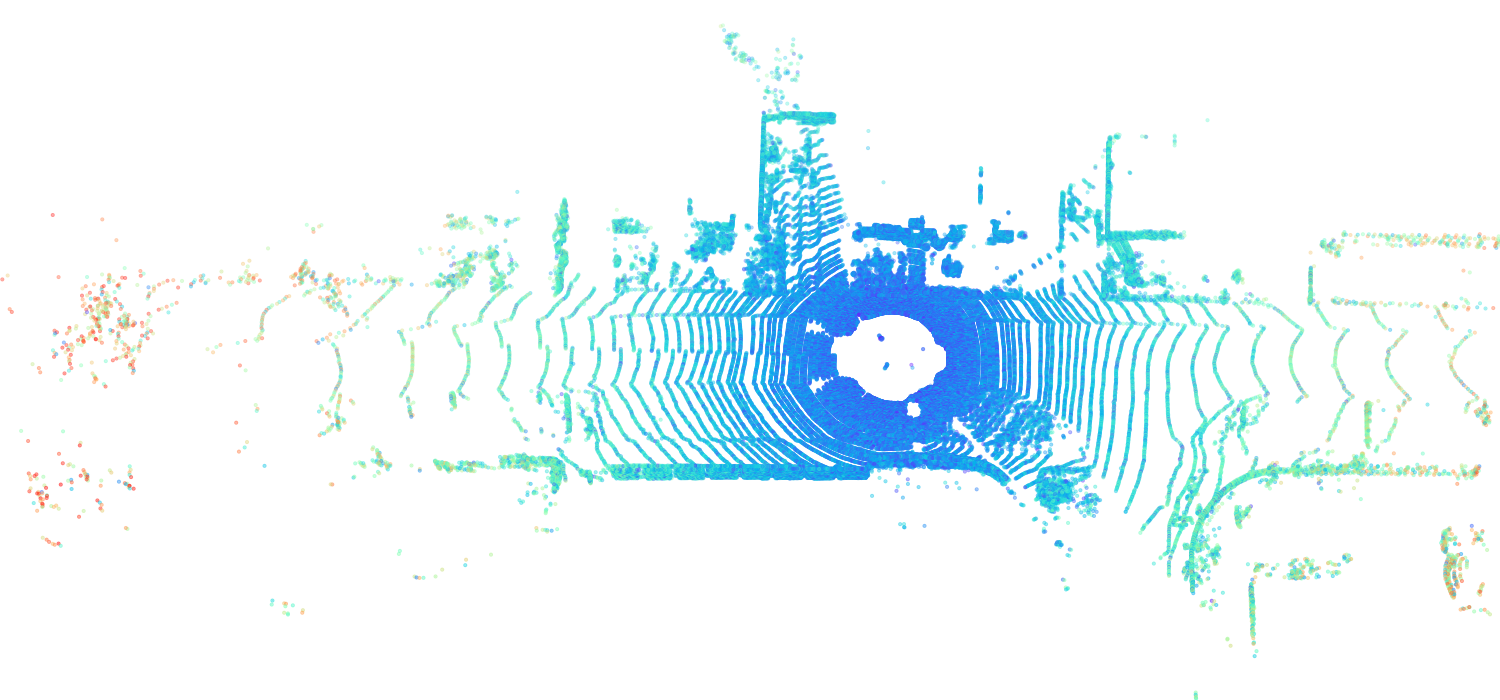}
    \label{fig:vis2_same}
    }

    \subfigure{
    \includegraphics[width=0.99\textwidth]{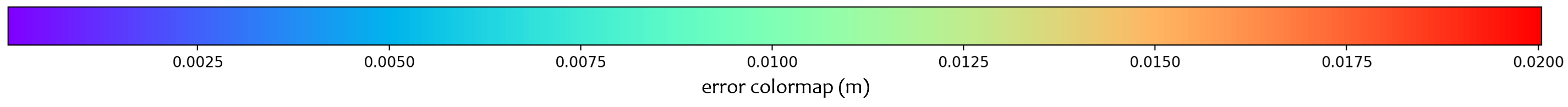}
    }
    \caption{
    The figures compare the reconstruction errors on point clouds No. 000000 in Sequence 11 (a, b, c) and No. 000000 in Sequence 16 (d, e, f) from SemanticKITTI~\cite{kitti}, among the baseline EHEM~\cite{ehem} method in (a, d), our SCP-EHEM method in (b, e), and the SCP-EHEM without a multi-level Octree in (c, f). The metrics are D1 PSNR @ bpp. The error colormap is displayed below the figures, where purple indicates the lowest error, and red represents the highest error.
    It is evident that the central parts of the point clouds in (b, e) have significantly lower reconstruction errors than the baseline method shown in (a, d). Conversely, the distant parts in (c, f) have much higher reconstruction errors than the central parts, but this is mitigated by implementing the multi-level Octree as depicted in (b, e).}
    \label{fig:error}
    \vspace{-4mm}
\end{figure*}

We evaluate point cloud compression methods across two dimensions: rate and distortion. The rate is measured in bits per point (bpp), which is the quotient of the bitstream length and the number of points. For distortion, we employ the most frequently used three metrics: point-to-point PSNR (D1 PSNR), point-to-plane PSNR (D2 PSNR)~\cite{d2psnr}, and Chamfer distance (CD). These metrics are adopted in the former methods, such as EHEM~\cite{ehem} and OctAttention~\cite{octattention}, making them convenient to compare with former methods.
% so it is efficient to compare SCP with the former methods on these metrics.
We also employ BD-Rate~\cite{bdrate} to illustrate our experimental results clearly. Following EHEM~\cite{ehem} and \cite{muscle}, we set the peak value of PSNR to 59.70 for SemanticKITTI and 30000 for Ford.

\subsubsection{Implementation Details}

Our experiments are conducted on OctAttention~\cite{octattention} and our reproduced EHEM~\cite{ehem}, as the official implementation of EHEM is not provided. During training, we input the information of Spherical-coordinate-based Octree to the models, including the following information:
% \begin{itemize}
(1) Octant: the integer order of the current voxel in the parent voxel, in the range of $[1, 8]$;
(2) Level: the integer depth of the current voxel in the Octree, in the range of $[1, 16]$ and $[1, 17]$ for SemanticKITTI and Ford, respectively;
(3) Occupancy, Octant, Level of ancient voxels: All the information of former levels. We trace back 3 levels, the same as OctAttention~\cite{octattention};
(4) Position: three floating-point positions of the current voxel in the Spherical coordinates, regularized to range $[0, 1]$.
% \end{itemize}

For the multi-level Octree method, we can deduce from \eref{eq:error_c} and \eref{eq:error_s} that $\varepsilon_s = \varepsilon_c,\ 2\varepsilon_c,\ 4\varepsilon_c$ at approximately $\rho=0.247\rho_{max},\ 0.493\rho_{max},\ 0.986\rho_{max}$, respectively. On the other hand, SCP predicts the distribution of each part separately, which means we repeat the processing of the first several levels $N$ times. Hence, for efficiency, $t_n$ is set to the inverse of the power of 2, which utmost reduces repetition. In our experiments, to keep the reconstruction error of SCP lower than that in Cartesian coordinates, we set $N=3$ and $t_n\in\{0, \frac{1}{4}, \frac{1}{2}\}$, which divides the point clouds into 3 parts: the inner part with $\rho\in[0, \frac{1}{4}\rho_{max})$, the middle part with $\rho\in[\frac{1}{4}\rho_{max}, \frac{1}{2}\rho_{max})$, and outer part with $\rho\in[\frac{1}{2}\rho_{max}, \rho_{max}]$. This splitting method constrains the upper bound of the reconstruction error to be lower or equal to the one in Cartesian coordinates for most areas, thereby achieving similar or better reconstruction results for both distant and central parts in LiDAR PCs.

We train SCP models on SemanticKITTI and Ford datasets for 20 and 150 epochs, respectively. We employ the default Adam optimizer~\cite{adam} for all experiments with a learning rate of $10^{-4}$. All the experiments are done on a server with 8 NVIDIA A100 GPUs and 2 AMD EPYC 7742 64-core CPUs.
\vspace{-1mm}

\subsubsection{Baselines}
We use recent methods as our baseline to verify the effectiveness of our SCP method. These methods include the state-of-the-art EHEM~\cite{ehem}, which employs the Swin Transformer~\cite{swin} and checkerboard~\cite{checkerboard} to compress point clouds; the sparse-CNN-based method SparsePCGC~\cite{nju}, which achieves high compression performance with an end-to-end scheme; the aforementioned OctAttention~\cite{octattention}, which uses the Transformer and sibling voxel contexts; and the MPEG G-PCC TMC13~\cite{gpcc} with either Octree or predictive geometry, which is the mainstream hand-crafted compression method. Note that the predictive geometry is only available for the Ford dataset since the SemanticKITTI dataset lacks the necessary sensor information for the calculation of predictive geometry.

\begin{table}[]
\centering
\begin{tabular}{c|c|c|c}
\toprule
Method   & Depth=12 & Depth=14 & Depth=16 \\
\midrule
G-PCC     &   0.22 / 0.06  & 0.63 / 0.21 & 1.05 / 0.39 \\\hline
EHEM     &   0.60 / 0.57     &   1.48 / 1.57     &  2.92 / 3.13   \\\hline
\makecell[c]{SCP-E w/o} &   0.63 / 0.62     &    1.73 / 1.86    &   3.21 / 3.49    \\\hline
SCP-EHEM &   1.16 / 1.11     &    2.34 / 2.41    &     3.92 / 4.09\\\bottomrule
\end{tabular}
\caption{Encoding/decoding times (in seconds) for a D-depth octree on SemanticKITTI dataset among GPCC~\cite{gpcc}, EHEM~\cite{ehem}, SCP-EHEM without multi-level Octree (SCP-E w/o), and SCP-EHEM. Runtimes for G-PCC are total times.}
\label{tab:time}
\vspace{-5mm}
\end{table}

\subsection{Experimental Results}
\label{sec:exp_res}

We designate SCP with a multi-level Octree as the default method, referred to as ``SCP'', while the SCP method without a multi-level Octree is called ``SCP w/o multi-level'' or ``SCP w/o''. The SCP methods with EHEM~\cite{ehem} and OctAttention~\cite{octattention} as the backbone are represented as ``SCP-EHEM'' and ``SCP-OctAttn'', respectively.
\vspace{-4mm}

\subsubsection{Quantitative results}
As depicted in \fref{fig:res}, our SCP-EHEM models achieve significant improvement when compared with all the baselines. SCP achieves a 61.23\% D1 PSNR BD-Rate gain on the SemanticKITTI dataset and 72.54\% on the Ford dataset when compared with Octree-based G-PCC. On the other hand, SCP-EHEM and SCP-OctAttention models outperform the baseline methods by 13.02\% and 29.14\% D1 PSNR BD-rate improvement on the SemanticKITTI dataset, respectively; 25.54\% and 24.78\% on the Ford dataset, respectively.
Remarkably, SCP-OctAttention even surpasses the original EHEM method, which has an 8 times larger receptive field than OctAttention. Our results on D2 PSNR and CD metrics also significantly outperform the previous method, demonstrating the robustness of SCP in efficiently compressing various point clouds under different driving environments for point cloud acquisition.
For the results of predictive-geometry-based G-PCC on the Ford dataset, it has a much better performance than the Octree-based one except for the lowest reconstruction quality. However, it is surpassed by both SCP-EHEM and SCP-OctAttention because of the lack of context information.
All these experiments showcase the universality of our SCP, which can be applied to various learned methods to fully leverage the circular feature of LiDAR PCs and improve the performance of these methods.

We show the inference time comparisons among G-PCC, the reproduced EHEM, SCP-EHEM without multi-level Octree (SCP-E w/o), and SCP-EHEM in \tref{tab:time}. The reconstruction performance of SCP-E w/o is better than the original Cartesian-coordinate-based EHEM because Spherical-coordinate-based Octree has lower reconstruction error in the central parts, which contain most points in the point cloud. This reserves more individual points after quantization, leading to a correspondingly longer processing time. On the other hand, one limitation of SCP-EHEM is that it repeats calculations of the beginning levels for each part, resulting in time overhead, which can be eliminated by processing the former levels together for all 3 parts. We will solve the problem of this overhead in our future work.
\vspace{-1mm}

\begin{figure}[h!]
    \centering
    \subfigure[Ablations on SCP-EHEM.]{
    \includegraphics[width=0.22\textwidth]{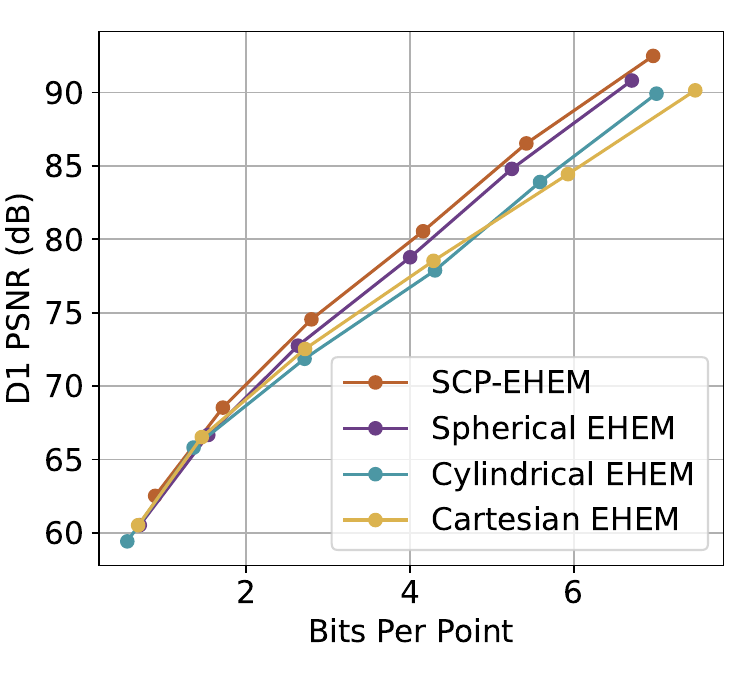}
    }
    \subfigure[Ablations on SCP-OctAttn.]{
    \includegraphics[width=0.22\textwidth]{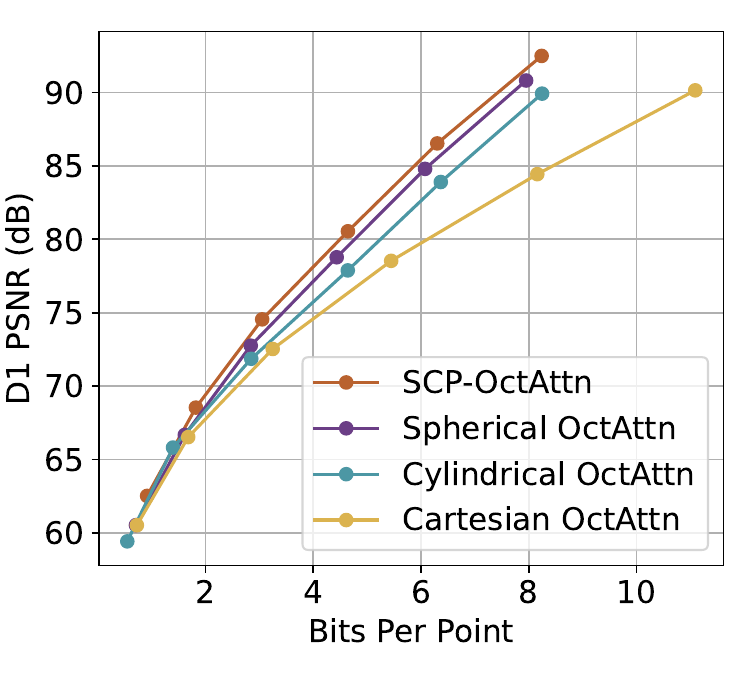}
    }
    \vspace{-2mm}
    \caption{Ablation experiments among our SCP, the Spherical-coordinate-based methods (SCP without multi-level Octree method), the Cylindrical-coordinate-based methods, and the Cartesian-coordinate-based methods(original EHEM and OctAttn). The BD-Rate gains between SCP-EHEM and the others are 5.98\%, 14.61\%, and 13.02\%; while the ones between SCP-OctAttn and the others are 6.07\%, 12.70\%, 29.14\%, respectively.}
    \label{fig:abl}
    \vspace{-3mm}
\end{figure}

\subsubsection{Qualitative results}
As depicted in \fref{fig:error}, we visualize the reconstruction errors for point clouds in the SemanticKITTI dataset. The errors are represented with a rainbow color scheme, where purple indicates lower error and red signifies higher error. It is evident that the central parts of \fref{fig:vis_mul} and \fref{fig:vis2_mul} have lower errors than those of \fref{fig:vis_cart} and \fref{fig:vis2_cart}, while the distant parts of our method exhibit similar reconstruction errors. In another perspective, the distant area of \fref{fig:vis_same} and \fref{fig:vis2_same} exhibit significantly higher errors compared to the central areas. This issue is substantially mitigated following the application of our multi-level Octree method, as demonstrated in \fref{fig:vis_mul} and \fref{fig:vis2_mul}.
\vspace{-2mm}

\subsection{Ablation Study}

To validate the effectiveness of SCP and the multi-level Octree method, we conducted ablation studies on them, as shown in \fref{fig:abl}. For the selection of metric, we find the curve tendencies of 3 metrics in \fref{fig:res} are similar, so we follow the ablation settings of EHEM, using D1 PSNR as the metric.

To verify the good performance of SCP, we experiment on all three coordinate systems, Spherical, Cylindrical, and Cartesian, with both backbone models EHEM and OctAttention. Note that the Cartesian-coordinate-based methods are the original EHEM/OctAttention. These experiments are conducted without the multi-level Octree method because it is designed for angle-based coordinate systems and is not applicable to Cartesian-coordinate-based Octree. As shown in \fref{fig:abl}, Spherical-coordinate-based methods (Spherical ones) are better than both the Cartesian-coordinate-based ones (Cartesian methods) and the Cylindrical-coordinate-based ones (Cylindrical methods).
As mentioned in \fref{fig:split}, the Octree under Spherical coordinates has better relevant context for each point in the point cloud than that under Cylindrical/Cartesian coordinates. The improvement is introduced by the azimuthal coordinate and polar ($\rho$, $\theta$) coordinates in the Spherical coordinate system.
For the ablation study on the multi-level Octree method, our SCP methods outperform the version without our multi-level Octree method (SCP-$*$ w/o), proving the effectiveness of our method.

\vspace{-1mm}

\section{Conclusion}
\label{sec:conclu}

We introduce a model-agnostic Spherical-Coordinate-based learned Point cloud compression (SCP) method that effectively leverages the circular shapes and azimuthal angle invariance feature present in spinning LiDAR point clouds. Concurrently, we address the higher reconstruction error issue in distant areas of the Spherical-coordinate-based Octree by implementing a multi-level Octree method. Our techniques exhibit high universality, making them applicable to a wide range of learned point cloud compression methods. Experimental results on two backbone methods, EHEM and OctAttention, exhibit the high effectiveness of our SCP and multi-level Octree method. On the other hand, our methods still have the potential to be improved on the inference time, which will be done in our future work.

\newpage
\section*{Acknowledgments}

These research results were obtained from the commissioned research (No. JPJ012368C06801, JPJ012368C03801) by National Institute of Information and Communications Technology (NICT), Japan; JSPS KAKENHI Grant Number JP23K16861.

\bibliography{aaai24}

\newpage
\section*{Appendix}
% \section{Predictive Geometry}

% \section{Cylindrical and Spherical}

\subsection*{Reconstruction error of Cartesian coordinates and Spherical coordinates}
\label{sec:apdx_eq}

We illustrate the derivation of \eref{eq:error_c} and \eref{eq:error_s} in this section. We use the $L2$-norm to evaluate the reconstruction error. This $L2$-norm is used in all three metrics in our experiments.

For the Octree in Cartesian coordinates, the leaf voxels are cubes whose side length is quantization step $q$. The position of a voxel is the center of it, marked as $(x_c, y_c, z_c)$. The upper bound of reconstruction error can be reached at the corners of the cube, one of which can be written as $(x_c+\frac{q}{2}, y_c+\frac{q}{2}, z_c+\frac{q}{2})$. Therefore, the upper bound of reconstruction error in Cartesian coordinates is

\begin{equation}
\begin{aligned}
    \varepsilon_c\leq\sqrt{3(\frac{q}{2})^2} = \frac{\sqrt{3}q}{2}.
\end{aligned}
\end{equation}

\begin{figure}[h!]
    \centering
    \includegraphics[width=0.45\textwidth]{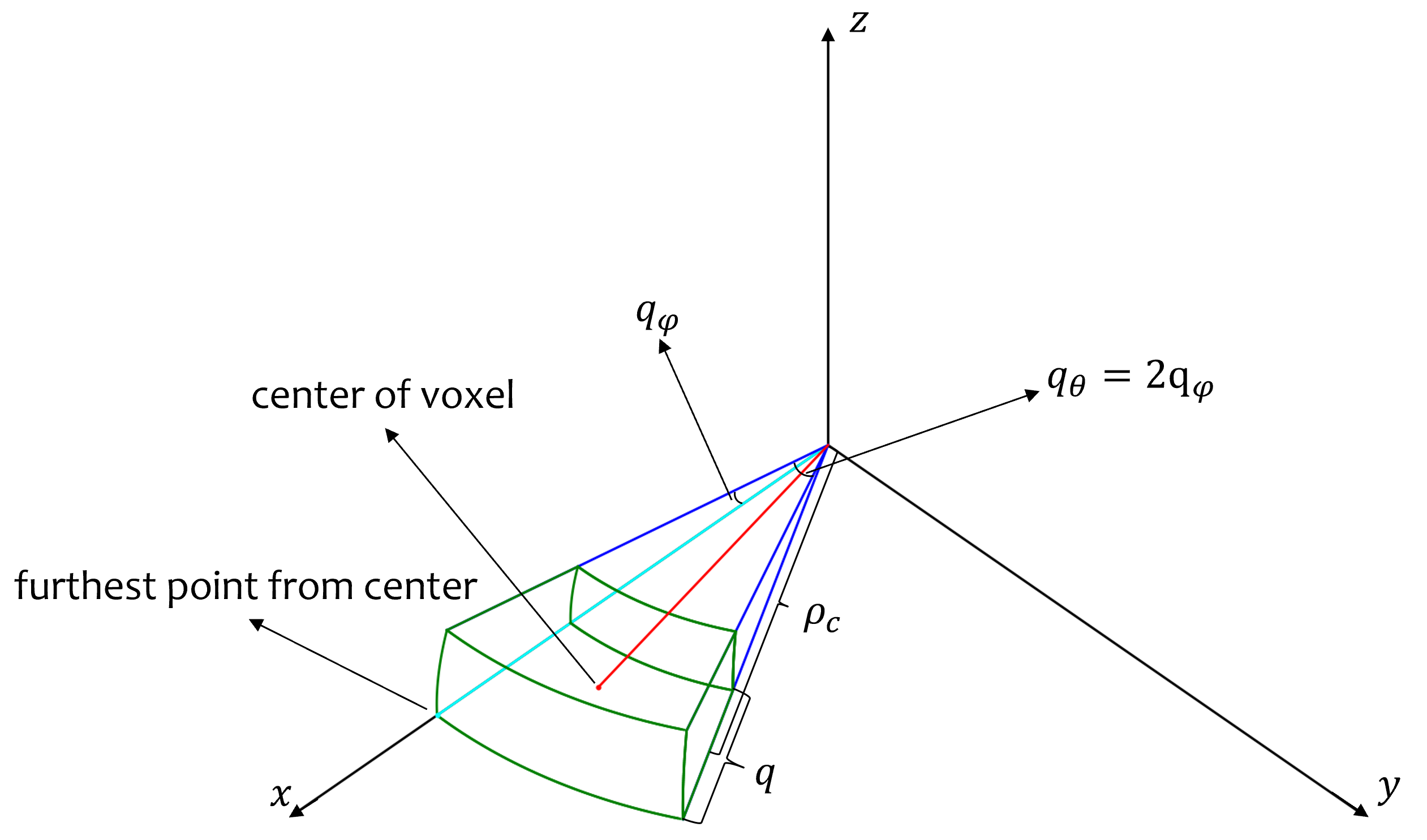}
    \caption{Derivation for \eref{eq:error_s}.}
    \label{fig:eq6}
\end{figure}

To compare with former methods, we need to convert Spherical coordinates back to Cartesian coordinates for reconstruction error calculation. In Spherical coordinates, the size of voxels is only relevant to coordinate $\rho$, therefore, we analyze the one voxel beside coordinate $x$, as the green box shown in \fref{fig:eq6}, which is easier for derivation. The voxel center can be marked as $(\rho_c\cos\frac{q_\theta}{2}, \rho_c\sin\frac{q_\theta}{2}, \rho_c\sin\frac{q_\theta}{4})$. We use $q_\theta$ to replace $q_\varphi$ according to \eref{eq:qs}. In the most situation, $q$ is much smaller that $\rho_c$, so we can set $\cos\frac{q_\theta}{2}=1$, $\sin\frac{q_\theta}{2}=\sin\frac{q_\theta}{4}=1$. Then the position can be rewritten as: $(\rho, \rho\frac{q_\theta}{2}, \rho\frac{q_\theta}{4})$.
On the other hand, one of the furthest points in the voxel is indicated in \fref{fig:eq6}. The position of it is $(\rho_c + \frac{q}{2}, 0, 0)$, which can be approximated as $(\rho_c, 0, 0)$ since $q$ is much smaller than $\rho_c$.
In conclusion, the upper bound of the reconstruction error can be derived by calculating the distance between these two points, as shown in \eref{eq:eq6}.

\begin{equation}
\begin{aligned}
    \varepsilon_s \leq \sqrt{(\rho_c\frac{q_\theta}{2})^2+(\rho_c\frac{q_\theta}{4})^2} = \frac{\sqrt{5}}{4}\rho_c q_\theta = \frac{\sqrt{5}\pi q}{2\rho_{max}}\rho_c.
    \label{eq:eq6}
\end{aligned}
\end{equation}

% \subsection{Inference Time Comparisons}

% We show the inference time comparisons among the reproduced EHEM, SCP-EHEM without multi-level Octree, and SCP-EHEM in \tref{tab:time}. Note that our reconstruction error is lower than the original EHEM at each depth because there are more voxels in the Spherical-coordinate-based Octree, leading to longer coding time for SCP-EHEM without multi-level Octree. In the meanwhile, SCP-EHEM repeats calculating the beginning levels for each part, resulting in an overhead. This overhead occupies a higher ratio in shallow Octree and gets a lower ratio in deeper ones.

% \begin{table}[]
% \caption{Inference times (in seconds) for encoding/decoding a D-depth octree on KITTI dataset among EHEM~\cite{ehem}, SCP-EHEM without multi-level Octree, and SCP-EHEM.}
% \label{tab:time}
% \centering
% \begin{tabular}{c|c|c|c}
% \toprule
% Method   & Depth=12 & Depth=14 & Depth=16 \\
% \midrule
% GPCC     &   0.22 / 0.06  & 0.63 / 0.21 & 1.05 / 0.39 \\\hline
% EHEM     &   0.60 / 0.57     &   1.48 / 1.57     &  2.92 / 3.13   \\\hline
% \makecell[c]{SCP-E w/o} &   0.63 / 0.62     &    1.73 / 1.86    &   3.21 / 3.49    \\\hline
% SCP-EHEM &   1.16 / 1.11     &    2.34 / 2.41    &     3.92 / 4.09\\\bottomrule
% \end{tabular}
% \end{table}

\end{document}